\documentclass[10pt,twocolumn,twoside]{IEEEtran}
\usepackage{multirow}
\usepackage{booktabs}
\usepackage{graphicx}
\usepackage{color}
\usepackage{url}
\usepackage[colorlinks,
            urlcolor=black,
            linkcolor=black,
            anchorcolor=black,
            citecolor=black
            ]{hyperref}

\usepackage{cite}

\ifCLASSINFOpdf
\else
\fi
\usepackage[cmex10]{amsmath}

\begin{document}
%
\title{Nighttime Haze Removal with Illumination Correction}
%
%
%

\author{Jing Zhang~\IEEEmembership{Student Member,~IEEE}, Yang Cao~\IEEEmembership{Member,~IEEE}, 
Zengfu Wang~\IEEEmembership{Member,~IEEE}

\thanks{Jing Zhang, Yang Cao and Zengfu Wang are with the Department of Automation, University of Science and Technology of China, Hefei, P.R.China. (e-mail: \{forrest, zfwang\}@ustc.edu.cn)}
}

\maketitle

\begin{abstract}
Haze removal is important for computational photography and computer vision applications. However, most of the existing methods for dehazing are designed for daytime images, and cannot always work well in the nighttime. Different from the imaging conditions in the daytime, images captured in nighttime haze condition may suffer from non-uniform illumination due to artificial light sources, which exhibit low brightness/contrast and color distortion. In this paper, we present a new nighttime hazy imaging model that takes into account both the non-uniform illumination from artificial light sources and the scattering and attenuation effects of haze. Accordingly, we propose an efficient dehazing algorithm for nighttime hazy images. The proposed algorithm includes three sequential steps. i) It enhances the overall brightness by performing a gamma correction step after estimating the illumination from the original image. ii) Then it achieves a color-balance result by performing a color correction step after estimating the color characteristics of the incident light. iii) Finally, it remove the haze effect by applying the dark channel prior and estimating the point-wise environmental light based on the previous illumination-balance result. Experimental results show that the proposed algorithm can achieve illumination-balance and haze-free results with good color rendition ability.
\end{abstract}


%
\IEEEpeerreviewmaketitle

\section{Introduction}
\label{sec:intro}
%
%
%
%
\IEEEPARstart{H}{aze} may change the colors and reduce the contrast of the captured images. The degradation is mainly caused by the light scattering and light attenuation in the atmosphere. It is important to remove the haze from the degraded images for different applications.

Many dehazing methods have been proposed to deal with daytime haze images. Multiple-image based methods require two or more images of the same scene either under different atmospheric conditions (e.g., the dichromatic method proposed in \cite{CVPR_2000_Narasimhan, TPAMI_2003_Narasimhan}) or polarization states (e.g., the polarization based methods \cite{CVPR_2001_Schechner, CVPR_2006_Shwartz}) for turning the ill-posed problem into a well posed or over-constrained one. Since it is difficult to obtain the required images, many researchers have proposed some single image based haze removal methods by using different priors \cite{CVPR_2008_Tan, TOG_2008_Fattal, TPAMI_2011_He, RSE_1988_Chavez}. For instance, Chavez proposes a dark-object subtraction technique for atmospheric scattering correction of multispectral data in \cite{RSE_1988_Chavez}. It is assumed that there must be an object that is dark in each channel of the multispectral data. Moreover, since all the scene points have the same distance with the camera in satellite images, Chavez assumes that the transmission variables can be treated as a constant independent of the position. Based on the above assumptions, the number of transmission variables reduces to one and it can be obtained easily given the atmospheric light. However, as indicated in \cite{Phd_2011_He}, the transmission variables are indeed not a single constant in most natural haze images, where the scene depth is not constant. Therefore, the constant-transmission assumption limits its application in special cases, such as satellite images. Partly inspired by the dark object subtraction technique, He et al. propose the well-known dark channel prior and the corresponding effective single image dehazing method \cite{Phd_2011_He, TPAMI_2011_He}, which can obtain fairly good result with low computational cost \cite{Phd_2011_He, ECCV_2010_He}. In addition, Tan proposes a visibility maximization method to enhance the visibility of the haze image under the constraint of the haze imaging model \cite{CVPR_2008_Tan}, and Fattal solves the haze removal problem by Independent Component Analysis (ICA) \cite{TOG_2008_Fattal}. We recommend \cite{Phd_2011_He} for more details and analysis. Very recently, Tang et al. propose a new image dehazing algorithm by investigating haze-relevant features in a learning framework \cite{CVPR_2014_Tang}, where they use Random Forest to learn a regression model for estimating the transmission for hazy images. They show that the dark-channel feature is the most informative one for haze removal while other haze-relevant features also contribute significantly in a complementary way. Though they only use synthetic hazy image patches for training, their algorithm outperforms state-of-the-art methods.

\begin{figure}
\centering
\includegraphics[width=0.8\linewidth]{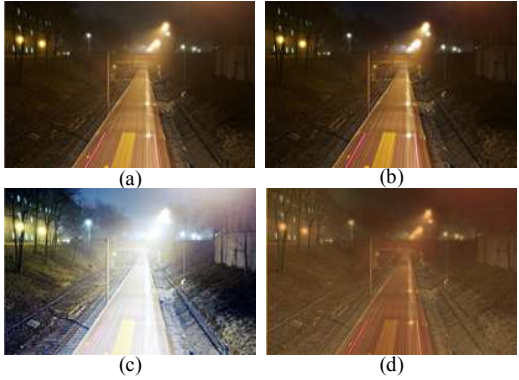}
\caption{(a) A nighttime haze image. (b) Dehazing reuslt of \cite{ECCV_2010_He}. (c) Histogram equalization result. (d) Result of \cite{ICIP_2012_Pei}.}
\label{fig:intro}
\end{figure}

However, these above dehazing methods are designed for daytime images and cannot always work well for nighttime haze images. Due to the different imaging conditions, e.g., non-uniform illumination from the artificial light sources, the nighttime dehazing problem is much more challenging than the daytime case. For instance, Fig.~\ref{fig:intro}(a) shows a nighttime haze image and Fig.~\ref{fig:intro}(b) shows the dehazing result of \cite{ECCV_2010_He}. It can be seen that it has little dehazing effect by using the dark channel prior directly, which has been indeed misused since the different properties of images from daytime/nighttime imaging conditions. Please find the high quality figures on the webpage\footnote{\href{http://staff.ustc.edu.cn/\%7Eforrest/NighttimeDehazing.htm}{{http://staff.ustc.edu.cn/\%7Eforrest/NighttimeDehazing.htm}}}. 

\begin{figure}
\centering
\includegraphics[width=0.8\linewidth]{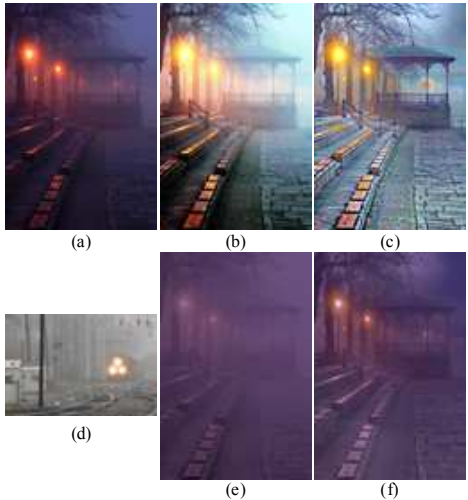}
\caption{(a) A nighttime haze image. (b) Histogram equalization result. (c) Result of the proposed algorithm. (d) A daytime haze image used as the target image in \cite{ICIP_2012_Pei}. (e) Statistic correction result of \cite{ICIP_2012_Pei}. (f) Final dehazing result of \cite{ICIP_2012_Pei}.}
\label{fig:icip_tran}
\end{figure}

To the best of our knowledge, there are little literatures about nighttime haze removal in the past decades. Histogram equalization (Heq) is one popular method to enhance the overall contrast of an image. Fig.~\ref{fig:intro}(c) shows the corresponding resultㄛwhich is more visual-pleasing than Fig.~\ref{fig:intro}(a)-(b). The contrasts of some regions like walls, trees and buildings have been enhanced. However, due to the unbalanced illumination at different points and the resulting wide gap of intensities between them, Heq amplified the intensities of regions which are bright in the original image, such as the light sources and the railway in Fig.~\ref{fig:intro}(c). Consequently, those regions are overexposure-like and lose some details. Fig.~\ref{fig:icip_tran}(b) shows another example.  Recently, Pei et al. propose a haze removal method for nighttime images \cite{ICIP_2012_Pei}. They firstly transfer the input nighttime haze image into a grayish one, and then apply a refined dark channel prior to remove the haze. Furthermore, to achieve better results they apply bilateral filter to perform a local contrast correction on the dehazing results. Their method can achieve result with more details than the original image (Fig.~\ref{fig:intro}(d) and Fig.~\ref{fig:icip_tran}(f)). However, since its color transfer procedure needs a given target image (usually a daytime haze image, e.g., Fig.~\ref{fig:icip_tran}(d)), it changes the color characteristics of the input nighttime haze image according to the target one, despite their different scene contents and imaging conditions. Consequently, the global color transfer procedure in the Lab color space usually leads to a complete grayish result (Fig.~\ref{fig:icip_tran}(e)). Such a result may be different from the expected illumination-balance one and will affect the final dehazed result (Fig.~\ref{fig:icip_tran}(f)). Fig.~\ref{fig:intro}(d) shows another example.

\begin{figure}
\centering
\includegraphics[width=0.85\linewidth]{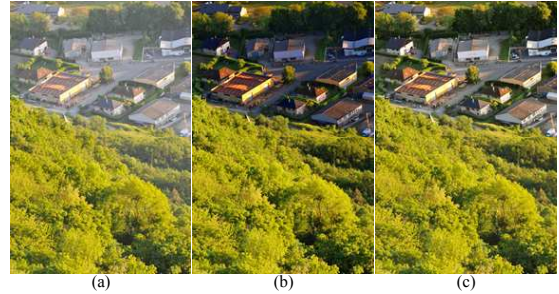}
\caption{(a) A daytime haze image. (b) Dehazing result of \cite{TPAMI_2011_He}. (c) Dehazing result of \cite{CVPR_2014_Tang}.}
\label{fig:daytimehaze}
\end{figure}

In \cite{ICIP_2014_Zhang}, we propose an efficient algorithm for nighttime haze removal that builds on a new imaging model. This new model takes into account both the non-uniform illumination from artificial light sources and the scattering and attenuation effects of haze. Based on this model, we give a novel dehazing algorithm including three sequential steps: illumination compensation, color correction and dehazing. First, it enhances the overall brightness by performing a gamma correction step after estimating the illumination from the original image. Then it achieves a color balance result by performing a color correction step after estimating the color characteristics of the incident light. Finally, it remove the haze effect by applying the dark channel prior and estimating the point-wise environmental light based on the previous illumination-balance result. In this paper, we extend that work from the following aspects: 1) polishing the new imaging model by including the derivation about the environmental light; 2) including more optimization details as well as the discussions about the relations between the above three components of the proposed algorithm; 3) conducting more experiments to evaluate the performance of the proposed algorithm, e.g., including an experiment on synthetic images to compare the effects by using gamma correction and polynomial fitting in the illumination correction step, and quantitatively evaluate the estimate of the incident light color as well as the final dehazing results. 4) presenting the analysis about its computational complexity and limitations.

\section{Related Work and Problem Analysis}
\label{sec:related_work}
For daytime haze image, e.g., an example as shown in Fig.~\ref{fig:daytimehaze}(a), it is often assumed that the haze is homogeneous and the only light source is the atmospheric light at infinity. For model-based methods \cite{CVPR_2000_Narasimhan, TPAMI_2003_Narasimhan, CVPR_2001_Schechner, CVPR_2006_Shwartz, CVPR_2008_Tan, TPAMI_2011_He}, the former one results a point-wise variable, i.e., transmission, which is identical for each channel and only depends on the scene depth. The latter one results a constant variable, i.e., atmospheric light, which is usually estimated from the pixel intensities of infinite atmospheric zone. In practice, the estimated atmospheric light is white or approximate white at least. Considering these two assumptions together, the dehazing algorithm indeed removes a white veil from the original haze image. Please see Fig.~\ref{fig:daytimehaze}(a) and Fig.~\ref{fig:daytimehaze}(b)-(c) for comparison.

\begin{figure}
\centering
\includegraphics[width=0.85\linewidth]{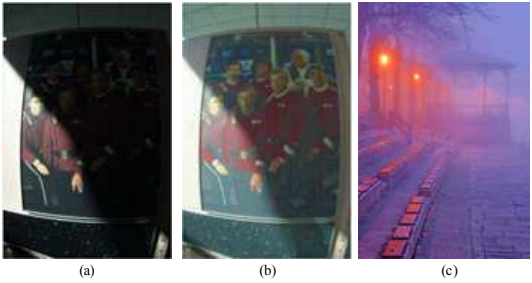}
\caption{(a) An illumination unbalanced image. (b) Result of (a) by using method in \cite{SBH_2005_Elad}. (c) Result of Fig.~\ref{fig:icip_tran}(a) by using method in \cite{SBH_2005_Elad}.}
\label{fig:retinex}
\end{figure}

\begin{figure*}
\centering
\includegraphics[width=0.85\linewidth]{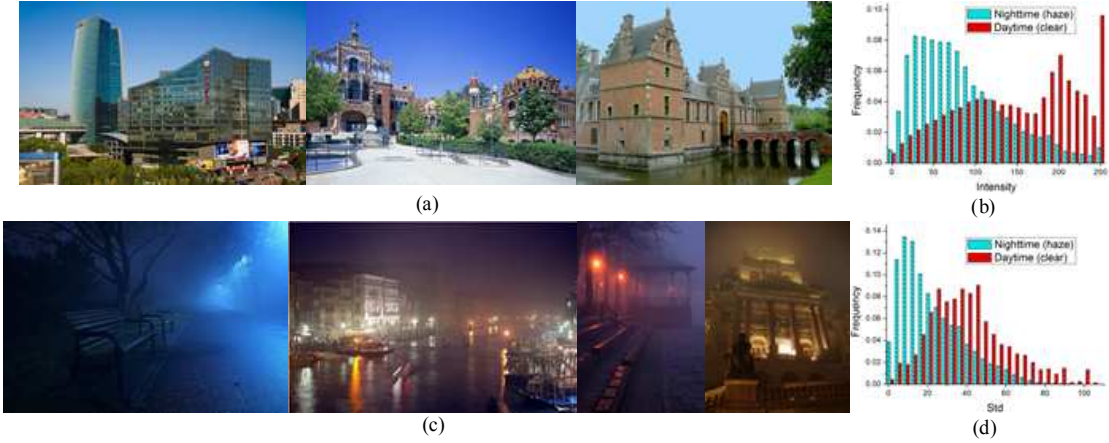}
\caption{(a) Some examples of clear daytime images. (b) Statistics of illumination intensities of clear and nighttime haze images. (c) Some examples of nighttime haze images.  (d) Statistics of standard deviations of values on local patches.}
\label{fig:illustration}
\end{figure*}

In literatures about retinex \cite{JVCIR_2003_Elad, SBH_2005_Elad, TIP_2014_Wang}, different methods have been proposed to enhance the image captured in non-uniform illumination environment. An image is treated as the product of reflectance and illumination. The former one depicts the intrinsic reflection property of object in a scene, and the latter one depicts the light intensity received by each scene point. Given an input image, estimating reflectance and illumination simultaneously is an ill-posed problem. In \cite{JVCIR_2003_Elad}, Elad et al. deal with it via the variational approach. By enforcing smoothness constraints on illumination and reflectance, Elad formulates it as an optimization problem and uses two bilateral filters to obtain the estimation results \cite{SBH_2005_Elad}. Wang et al. propose an variational Bayesian method for retinex by using Gibbs distributions as prior distributions for the reflectance image and the illumination image. With the Bayesian inference, the method can simultaneously estimates model parameters along with the unknown illumination image and reflectance image \cite{TIP_2014_Wang}. Fig.~\ref{fig:retinex}(a) shows an example image captured in non-uniform illumination environment. The light intensities in the left part are much higher than ones in the right part. It seems that the incident light is occluded when it goes to right part. Fig.~\ref{fig:retinex}(b) shows the enhanced results obtained by using method in \cite{SBH_2005_Elad}. Note that here the incident light often refers to the natural illumination, namely, white light. Thus no analysis on color characteristics of the incident light is involved. However, for nighttime haze environment, there are other influence factors, i.e., color characteristic of artificial light source which leads to color distortion of imaging objects, and the light scattering/attenuation of haze which reduce the contrast and visibility of a nighttime haze image. As shown in Fig.~\ref{fig:retinex}(c), the method in \cite{SBH_2005_Elad} balanced the intensities in the whole scene, but could not reduce the color distortion. Moreover, the haze effect is rather evident that some details of objects in the distance are still unavailable.

For Image captured in nighttime haze environment, i.e., low natural illumination and non-uniform artificial light sources condition, they exhibit some properties such as low overall brightness and non-uniform illumination. In addition, the haze may degrade the images's quality for its scattering and attenuation effects \cite{TPAMI_2011_He, Phd_2011_He}. Thus, images are generally in low contrast and lose some details. Moreover, the artificial light sources usually radiate color light which will be scattered by the haze. Consequently, it may also lead to color distortion of imaging objects. To give an illustration, we selected 20 illumination-balance images captured in the daytime which have lots of clear details and 20 haze images captured in the nighttime from Flickr. Some examples are shown in Fig.~\ref{fig:illustration}(a) and Fig.~\ref{fig:illustration}(c), respectively. Statistics of illumination values (values in V channels of images in HSV color space) are shown in Fig.~\ref{fig:illustration}(b). It can be seen the nighttime haze images have more low-intensity pixels than the clear ones. We also calculated the standard deviations of values on local patches. And the statistics of them are shown in Fig.~\ref{fig:illustration}(d). Images in the daytime shows higher variances which corresponds to their more clear details than haze images in the nighttime. For instance, we can see that the overall brightness of the leftmost image in Fig.~\ref{fig:illustration}(c) is very low, especially in the left part. The right part is a little brighter but dominated by the blue light. One can hardly distinguish the details of the trees and benches due to the influence of low illumination and scattering effect of haze in artificial light condition.

These influence factors mentioned above make the nighttime haze removal problem to be a very challenging one. Considering them together, the goal of this paper is to seek an algorithm which can solve the haze removal problem as well as the illumination-balance problem together. Fig.~\ref{fig:icip_tran}(c) shows the dehazing result by using the proposed algorithm on Fig.~\ref{fig:icip_tran}(a), which is illumination-balance and haze-free. It shows many clear details and is visual pleasing.

\section{A New Imaging Model For Nighttime haze environment}
\label{sec:model}

\subsection{Imaging model for daytime haze environment}
\label{subsec:daytime_model}
\begin{figure}
\centering
\includegraphics[width=0.9\linewidth]{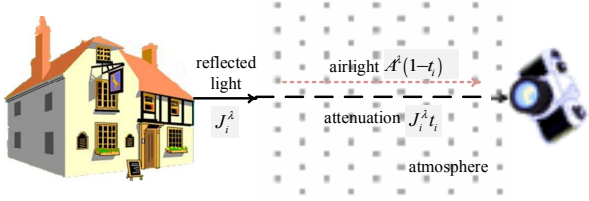}
\caption{A macro physical picture of the daytime haze imaging model. It is replotted from Fig.2.2 in \cite{Phd_2011_He}.}
\label{fig:daytime_hazemodel}
\end{figure}

Figure~\ref{fig:daytime_hazemodel} shows a macro physical picture of the daytime haze imaging model. The pixel intensity in the captured image consists of two parts: the direct attenuation part and the scattering part. The former one depicts that the light reflected from an object will be absorbed by the haze before it reaches at the camera. The latter one depicts that the haze scatters the light they absorb, playing as an infinite number of tiny light sources floating in the atmosphere \cite{Phd_2011_He}. Mathematically, it can be expressed as:

\begin{equation}
I_i^\lambda  = J_i^\lambda {t_i} + {A^\lambda }\left( {1 - {t_i}} \right) = {A^\lambda }R_i^\lambda {t_i} + {A^\lambda }\left( {1 - {t_i}} \right),
\label{eq:daytime_model}
\end{equation}
where $I_i^\lambda$ is the intensity of captured hazy image at location $i$ (we adopt a lexicographical order representation of an image.) and $\lambda$ represents one of the RGB channels. $J_i^\lambda$ is the expectation clear image (scene radiance). ${t_i}$ is the transmission at location $i$. ${A^\lambda }$ is the atmospheric light component of $\lambda$ channel. $R_i^\lambda$ is the reflectance which refers to the ratio of the reflected light to the illumination. It is related to the reflection characteristics of object surface.

The above model is adopted in the literatures of daytime image dehazing \cite{CVPR_2000_Narasimhan, TPAMI_2003_Narasimhan, CVPR_2001_Schechner, CVPR_2006_Shwartz, CVPR_2008_Tan, TPAMI_2011_He, CVPR_2014_Tang}, and shows its effectiveness for haze removal, especially when coupled with the dark channel prior \cite{TPAMI_2011_He}. However, this model is limited to daytime haze environment since it relies on the assumption that the atmospheric light is constant and the mainly light source in the scene. In nighttime haze environment, the light sources are mainly artificial light sources rather than the natural illumination, i.e., atmospheric light. Thus, the illumination as well as the scattered light will be affected by the particular locations and colors of artificial light sources.

\subsection{A new imaging model for nighttime haze environment}
\label{subsec:new_model}
\begin{figure}
\centering
\includegraphics[width=0.8\linewidth]{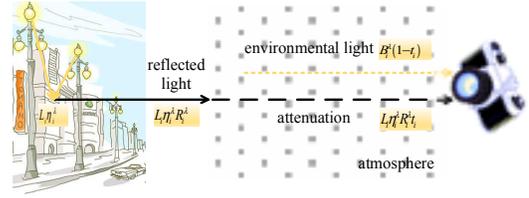}
\caption{A macro physical picture of the nighttime haze imaging model.}
\label{fig:nighttime_hazemodel}
\end{figure}

For nighttime haze removal problem, it has many similarities as well as differences with the daytime one. Figure~\ref{fig:nighttime_hazemodel} shows a macro physical picture of the nighttime haze imaging model. The captured image also consists of two terms: direct attenuation term and scattering light term. As analyzed in Section~\ref{subsec:daytime_model}, usually the atmospheric light is assumed to be the only light source for daytime haze environment, and the attenuation and scattering characteristics are identical for each channel, i.e., independent from the wavelength. However, in nighttime haze environment, both the magnitude and color of incident light have complex patterns since there are usually many artificial light sources with different colors and at different locations. It is difficult to give an explicit expression which depicts the changes of the incident light at different scene points. As a result, they are assumed to be local variables in this paper.

\begin{figure*}
\centering
\includegraphics[width=1\linewidth]{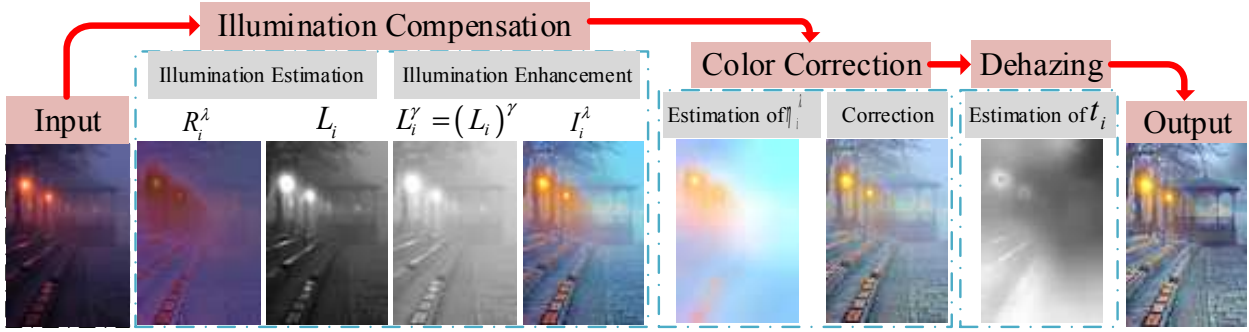}
\caption{A diagram of the proposed algorithm.}
\label{fig:flowchart}
\end{figure*}

Based on the above analysis, we propose a new imaging model for nighttime haze environment. Mathematically, it can be expressed as follows:
\begin{equation}
\begin{array}{c}
 I_i^\lambda  = {L_i}\eta_i^\lambda R_i^\lambda {t_i} + B_i^\lambda \left( {1 - {t_i}} \right) \\
 \qquad \buildrel \Delta \over = {L_i}\eta_i^\lambda R_i^\lambda {t_i} + {L_i}\sigma _i^\lambda \left( {1 - {t_i}} \right) \\
 \end{array}.
\label{eq:our_model}
\end{equation}
Here $I_i^\lambda$, $\lambda$, ${t_i}$ and $R_i^\lambda$ have the same meanings with the ones in Eq.~\eqref{eq:daytime_model}. ${L_i}$ is a scalar representing the illumination value, i.e., incident light intensity received by scene point located at $i$. $\eta_i^\lambda$ is a quantity accounts for the color characteristic of incident light. $B_i^\lambda $ represents the environmental light. It comes from two sources, i.e., incident light scattered around location $i$ and reflected light of objects scattered around location $i$. It can be formulated as follows:
\begin{equation}
\begin{array}{l}
 B_i^\lambda  = \frac{1}{{\left| \Omega  \right|}}\left( {\alpha \sum\limits_{j \in \Omega_i} {{L_j}\eta _j^\lambda }  + \left( {1 - \alpha } \right)\sum\limits_{j \in \Omega_i} {{L_j}\eta _j^\lambda R_j^\lambda } } \right) \\
 \quad \ \ \le \frac{1}{{\left| \Omega  \right|}}\left( {\alpha \sum\limits_{j \in \Omega_i} {{L_j}\eta _j^\lambda }  + \left( {1 - \alpha } \right)\sum\limits_{j \in \Omega_i} {{L_j}\eta _j^\lambda } } \right) \\
 \quad \ \  = \frac{1}{{\left| \Omega  \right|}}\sum\limits_{j \in \Omega_i} {{L_j}\eta _j^\lambda }  \\
 \end{array}.
\label{eq:environmental_light}
\end{equation}
where $\Omega_i$ represents the local neighborhood of position $i$, $\left| \Omega  \right|$ is the number of pixels in $\Omega_i$ and $\alpha $ is a weight, e.g., 0.5. The inequality holds since $R_i^\lambda$ lies in the range of [0,1]. We rewrite it as $B_i^\lambda  \buildrel \Delta \over = {L_i}\sigma _i^\lambda$ in Eq.~\eqref{eq:our_model} (Please see more details about $\sigma _i^\lambda$ in the Appendix). Different from the constant atmospheric light in Eq.~\eqref{eq:daytime_model}, it is replaced by a pointwise variable $B_i^\lambda $ in the new model to account for the intensity and color characteristic of environmental light.

Given the input hazy image, solving the variables in Eq.~\eqref{eq:our_model} is indeed an ill-posed problem. It seems more difficult than the daytime case since this new model involves more variables. In the following section, we propose an efficient algorithm to obtain the unknown variables based on simple assumptions. This algorithm consists of three sequential steps, i.e., illumination compensation, color correction and dehazing. A diagram is shown in Fig.~\ref{fig:flowchart}.

\section{A Sequential Algorithm for Nighttime Haze Removal}
\label{sec:solution}
The proposed algorithm to obtain the unknown variables relies on the following three assumptions: 1) the incident light received by each scene point is piecewise smooth; 2) the reflectance is piecewise continuous; 3) the transmission is piecewise smooth. These assumptions imply that the above local variables including $L_i$, $\eta_i^\lambda$, $\sigma _i^\lambda$ and ${t_i}$ are all piecewise smooth, and $R_i^\lambda$ is piecewise continuous. The first two assumptions are widely adopted in literatures abut retinex \cite{SBH_2005_Elad, TIP_2014_Wang}. The last assumption is also widely used in literatures about image dehazing \cite{TPAMI_2011_He, Phd_2011_He, CVPR_2014_Tang}.

\subsection{Illumination compensation}
\label{subsec:light_compensation}
Rewrite Eq.~\eqref{eq:our_model} as follows:

\begin{equation}
I_i^\lambda  = {L_i}\widehat{R_i^\lambda },
\label{eq:retinex}
\end{equation}
where $\widehat{R_i^\lambda } = \eta_i^\lambda R_i^\lambda {t_i} + \sigma _i^\lambda \left( {1 - {t_i}} \right)$. $\widehat{R_i^\lambda }$ is called surrogate reflectance in this paper. It is piecewise continuous (Please see the Appendix). Obtaining $L_i$ and $\widehat{R_i^\lambda}$ subjected to Eq.~\eqref{eq:retinex} is an ill-posed problem. Some approaches have been proposed in \cite{JVCIR_2003_Elad, SBH_2005_Elad, TIP_2014_Wang}. We use a similar energy formulation and optimization technique with \cite{SBH_2005_Elad} in this paper. A look-up-table log operation transfers the multiplication in Eq.~\eqref{eq:retinex} into an addition, resulting with $ii \buildrel \Delta \over = \log \left( I \right) = \log \left( L \right) + \log \left( {\widehat R} \right) \buildrel \Delta \over = ll + rr$. Note that we adopt a lexicographical order vectorization representation of $ii$, $ll$ and $rr$, and ignore the superscript $\lambda$ for simplicity. By enforcing smoothness constraints on $L$ and $\widehat{{R^\lambda }}$, the recovery of $ll$ and $rr$ can be formulated as the following optimization problem.

\begin{equation}
\begin{array}{c}
 \left\{ {ll,rr} \right\} = \mathop {\arg \min }\limits_{ll,rr:ll \ge ii} \left\{ {{\lambda _{ll}}\left\| {ll - ii} \right\|_2^2 + l{l^T}\Lambda ll} \right\} \\
 \qquad \qquad \qquad + \alpha \left\{ {{\lambda _{rr}}\left\| {rr - ii + ll} \right\|_2^2 + r{r^T}\Lambda rr} \right\} \\
 \end{array},
\label{eq:opt_retinex}
\end{equation}
where $\lambda _{ll}$, $\lambda _{rr}$, $\alpha$ are parameters, $\Lambda$ is the matting Laplacian matrix \cite{TPAMI_2008_Levin}, and the two term involves $\Lambda$ account for the smoothness penalty. The optimization problem can be efficiently solved (approximately) by using image guided filter \cite{ECCV_2010_He}. First, we transfer the input image $I$ into HSV color space, and use values in V channel as the initial estimate of $L$. Namely, we have an initial estimate of $ll$. Then, we apply an image guided filtering process on $ll$. Next, we can obtain an initial estimate of $rr$ by subtracting the filtering result of $ll$ from $ii$. Then, we apply an image guided filtering process on it to obtain the final estimate of $rr$. Finally, an exponential operation is applied to recover $L$ and $\widehat R$.

After obtaining the estimation of ${L_i}$ and $\widehat{R_i^\lambda }$, we apply a gamma correction to ${L_i}$ to balance the overall illumination of the image. Mathematically, it can be expressed as follows:
\begin{equation}
\widehat{I_i^\lambda } = {\left( {{L_i}} \right)^\gamma }\widehat{R_i^\lambda } = L_i^\gamma \eta_i^\lambda R_i^\lambda {t_i} + L_i^\gamma \sigma _i^\lambda \left( {1 - {t_i}} \right).
\label{eq:light_compensation}
\end{equation}
This nonlinear correction can enhance the illumination of dark regions while preventing bright regions from being amplified. This process is illustrated in ``illumination compensation'' part in Fig.~\ref{fig:flowchart}. There is an optional stretching operation after the illumination compensation step. Mathematically, it can be expressed as follows.

\begin{equation}
\widehat{I_i^\lambda }{\rm{ = }}\frac{{\widehat{I_i^\lambda }{\rm{ - }}\widehat{I_{\min }^\lambda }}}{{\widehat{I_{\max }^\lambda }{\rm{ - }}\widehat{I_{\min }^\lambda }}},
\label{eq:strech}
\end{equation}
where $\widehat{I_{\max }^\lambda }$ and $\widehat{I_{\min }^\lambda }$ are the maximum and minimum of intensities of $\widehat{{I^\lambda }}$ in channel $\lambda$. For robustness, in practice we selected the intensity values which rank at 95\% and 5\% as the maximum and minimum, respectively.

\subsection{Color correction}
\label{subsec:color_correction}
We can prove that the inequality $\sigma _i^\lambda \le \eta _i^\lambda $ approximately holds (Please see the Appendix). In addition, considering that $R_i^\lambda$ lies in the range of [0,1], we have:
\begin{equation}
\begin{array}{l}
 \widehat{I_i^\lambda } \le L_i^\gamma \eta _i^\lambda {t_i} + L_i^\gamma \sigma _i^\lambda \left( {1 - {t_i}} \right) \\
 \quad \ \le L_i^\gamma \eta _i^\lambda {t_i} + L_i^\gamma \eta _i^\lambda \left( {1 - {t_i}} \right) \\
 \quad \ \le L_i^\gamma \eta _i^\lambda  \\
 \end{array}.
\label{eq:inequality}
\end{equation}

Then we can obtain the lower bound of $\eta_i^\lambda$ as $\underline {\eta_i^\lambda }  = {{\widehat{I_i^\lambda }} \mathord{\left/
 {\vphantom {{\widehat{I_i^\lambda }} {L_i^\gamma }}} \right.
 \kern-\nulldelimiterspace} {L_i^\gamma }}$. More robustly, we calculate the maximum of each overlapped patch, and then average the overlapped ones to obtain to the raw estimate. It can be expressed as:

\begin{equation}
\underline {\eta_i^\lambda }  = \frac{{\mathop {\max }\limits_{j \in {\Omega _i}} \widehat{I_j^\lambda }}}{{\mathop {\max }\limits_{j \in {\Omega _i}} L_j^\gamma }},
\label{eq:inequality_Omega}
\end{equation}

Since $\eta_i^\lambda$ is piecewise smooth, the above raw estimate can be refined by enforcing smoothness constraint on it. The refinement procedure can be formulated as an quadratic optimization problem, which is similar to the refinement about the raw transmission map in \cite{TPAMI_2011_He, ECCV_2010_He}. Mathematically, it can be expressed as follows.

\begin{equation}
{\eta^\lambda } = \mathop {\arg \min }\limits_{{\eta^\lambda }} {\left\| {{\eta^\lambda } - \underline {{\eta^\lambda }} } \right\|^2} + \lambda {\left( {{\eta^\lambda }} \right)^T}\Lambda {\eta^\lambda },
\label{eq:color_optimization}
\end{equation}
where $\Lambda$ is the matting Laplacian matrix \cite{TPAMI_2008_Levin}, and the second term accounts for the smoothness penalty. The optimization problem can be efficiently solved (approximately) by using image guided filter \cite{ECCV_2010_He}. Since $\underline {\eta_i^\lambda }$ is the lower bound of the expected $\eta_i^\lambda$, we enhance the result of Eq.~\eqref{eq:color_optimization} by multiplying an amplifying factor. This amplifying factor is calculated according to the ratio ${{{{\left( {\frac{1}{3}\sum\nolimits_\lambda  {{\eta^\lambda }} } \right)}^{{\gamma _0}}}} \mathord{\left/
 {\vphantom {{{{\left( {\frac{1}{3}\sum\nolimits_\lambda  {{\eta^\lambda }} } \right)}^{{\gamma _0}}}} {\left( {\frac{1}{3}\sum\nolimits_\lambda  {{\eta^\lambda }} } \right)}}} \right.
 \kern-\nulldelimiterspace} {\left( {\frac{1}{3}\sum\nolimits_\lambda  {{\eta^\lambda }} } \right)}}$. The parameter ${\gamma _0}$ is set to 1/1.2 in this paper.

Divided by ${\eta_i^\lambda }$ in both side of Eq.~\eqref{eq:light_compensation}, it becomes:

\begin{equation}
\widetilde{I_i^\lambda } = {{\widehat{I_i^\lambda }} \mathord{\left/
 {\vphantom {{\widehat{I_i^\lambda }} {\eta_i^\lambda }}} \right.
 \kern-\nulldelimiterspace} {\eta_i^\lambda }}.
\label{eq:color_correction}
\end{equation}
$\widetilde{I_i^\lambda }$ in the left side represents the color correction result. After this step, the dominant environmental light is removed. This process is illustrated in ``color correction'' part in Fig.~\ref{fig:flowchart}.

\subsection{Dehazing}
\label{subsec:dehazing}

Following Eq.~\eqref{eq:color_correction}, we have:
\begin{equation}
\begin{array}{l}
 \widetilde{I_i^\lambda } = L_i^\gamma R_i^\lambda {t_i} + L_i^\gamma \left( {\frac{{\sigma _i^\lambda }}{{\eta _i^\lambda }}} \right)\left( {1 - {t_i}} \right) \\
 \quad \ \buildrel \Delta \over = L_i^\gamma R_i^\lambda {t_i} + L_i^\gamma \Delta \sigma _i^\lambda \left( {1 - {t_i}} \right) \\
 \end{array},
\label{eq:dehazing_pre}
\end{equation}
where $\Delta \sigma _i^\lambda \buildrel \Delta \over = {{\sigma _i^\lambda } \mathord{\left/
 {\vphantom {{\sigma _i^\lambda } {\eta _i^\lambda }}} \right.
 \kern-\nulldelimiterspace} {\eta _i^\lambda }}$. This term accounts for the residual color effect of the environmental light in the light correction result. Adopting the following notation, we can rewrite Eq.~\eqref{eq:dehazing_pre} as:

\begin{equation}
J_i^\lambda  \buildrel \Delta \over = L_i^\gamma R_i^\lambda,
\label{eq:J}
\end{equation}

\begin{equation}
A_i^\lambda  \buildrel \Delta \over = L_i^\gamma \Delta \sigma _i^\lambda,
\label{eq:A}
\end{equation}

\begin{equation}
\widetilde{I_i^\lambda } \buildrel \Delta \over = J_i^\lambda {t_i} + A_i^\lambda \left( {1 - {t_i}} \right),
\label{eq:dehazing}
\end{equation}
where $J_i^\lambda$ represents the expected clear image, and $A_i^\lambda$ is the environmental light. Eq.~\eqref{eq:dehazing} is similar to Eq.~\eqref{eq:daytime_model}, which only differs in the environmental light (atmospheric light ${A^\lambda}$ in Eq.~\eqref{eq:daytime_model}) in the scattering term. The estimation method of ${t_i}$ and $A_i^\lambda$ is similar to method in \cite{TPAMI_2011_He, Phd_2011_He} by using dark channel prior. However, since $A_i^\lambda$ is a local variable rather than a global constant in Eq.~\eqref{eq:daytime_model}, we estimate it in a local neighborhood rather than the whole image as described in \cite{TPAMI_2011_He}. This process is illustrated in ``dehazing'' part in Fig.~\ref{fig:flowchart}. And the final output is also shown in Fig.~\ref{fig:flowchart}. It can be seen that the proposed algorithm can obtain a haze-free and illumination-balance image.

Note that the above three components are all necessary and the order of them should not be exchanged. Without the illumination compensation step, the illumination will keep unbalanced and details in dark regions are still not clear. In addition, the low illumination will lead an unstable and unreliable estimate of $\eta _i^\lambda$ in Eq.~\eqref{eq:inequality_Omega}. So the ``illumination compensation'' step is necessary and should be ahead of the ``color correction'' step. As we know that the dark channel prior is a statistical observation on daytime clear images \cite{TPAMI_2011_He}. By using it to estimate the transmission, the hazy image is required to be illumination-balance and the dominant environmental light is the white atmospheric light (otherwise, the dark channel will be affected by the low illumination and the dominant environmental light color). So the ``color correction'' step should be ahead of the ``dehazing'' step.

\section{Experiments}
\label{sec:experiments}
To demonstrate the effectiveness of the proposed algorithm, we collected 20 haze images captured in nighttime environment from Flickr and formed a test dataset. Some examples can be found in Fig.~\ref{fig:illustration}(c). The performance of the proposed algorithm is evaluated both objectively and subjectively by utilizing this test dataset and test images about a color set. It is compared with related methods \cite{SBH_2005_Elad, TPAMI_2011_He, ECCV_2010_He, ICIP_2012_Pei} from the follwing aspects, i.e., light correction, image dehazing, statistical visual measure and color rendition. Besides, we synthesized the nighttime hazy images based on Middlebury 2005 dataset, and quantitatively evaluated the performance of the proposed algorithm. The analysis on its computational complexity and discussions about its limitation are presented at the end of this section.

\begin{figure*}
\centering
\includegraphics[width=0.85\linewidth]{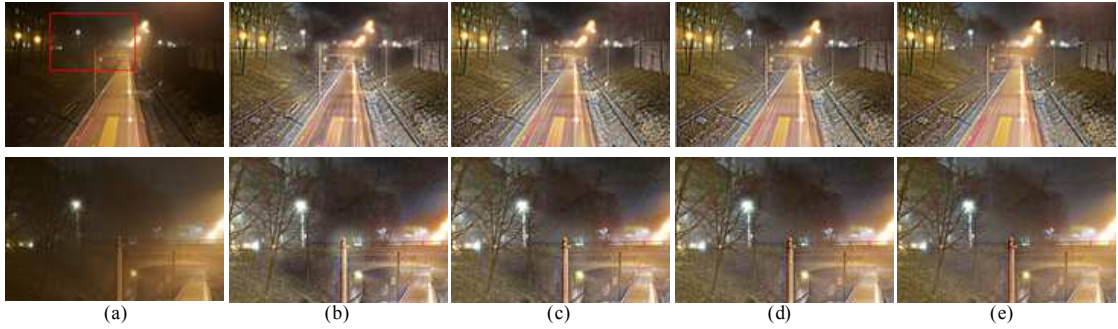}
\caption{Comparisons on different settings of kernel size of image guided filter. (a) Input nighttime haze image and the close-up view of the redbox region. (b)-(e) show the results of the proposed algorithm and the close-up views by setting different kernel sizes of image guided filter. The kernel radius of image guided filter is set to 8, 16, 32 and 64 in (b)-(e), respectively.}
\label{fig:parameter_gf}
\end{figure*}

\begin{figure*}
\centering
\includegraphics[width=0.85\linewidth]{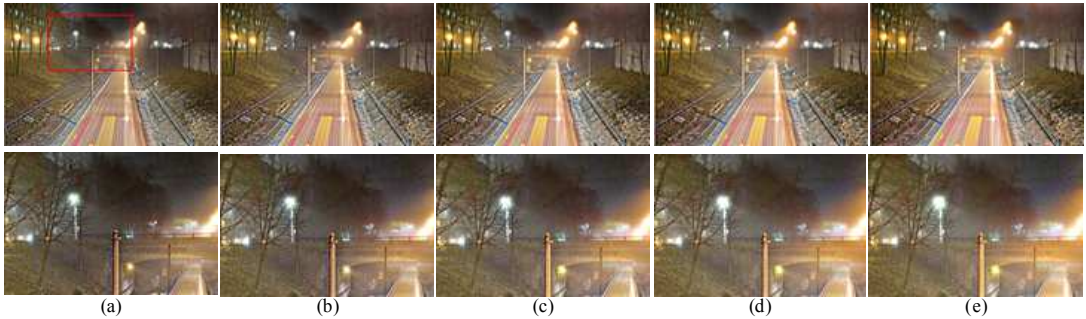}
\caption{Comparisons on different settings of local patch sizes when estimating $\eta_i^\lambda$ and $A_i^\lambda$. (a)-(e) show the results of the proposed algorithm and the close-up views by setting different local patch sizes. The radius of local patch is set to 3, 5, 7, 9 and 11 in (a)-(e), respectively.}
\label{fig:parameter_rs}
\end{figure*}

\subsection{Parameter settings}
\label{subsec:parameter_settings}
The parameter $\gamma$ is set to 1/3 in the illumination compensation part. The radius of local patch size is set to 5 when estimating $\eta_i^\lambda$ and $A_i^\lambda$. The kernel radius and the regularization parameter in image guided filter are set to 32 and 0.01, respectively. Note that the maximum size (either width or height) of all the images in the test dataset are around 500 pixels.

Figure~\ref{fig:parameter_gf} shows some results by setting different kernel sizes in image guided filter for the proposed algorithm. Other parameters are set as above accordingly. The kernel radiuses are 8, 16, 32 and 64 for Fig.~\ref{fig:parameter_gf}(b)-(e), respectively. It can be seen that there are some haze residuals in Fig.~\ref{fig:parameter_gf}(b)-(c), e.g., regions around the pillar and trees. The image guided filter is usually used as one-iteration approximate solution to the linear system for finding the matting result. As proved in \cite{Phd_2011_He}, using a large kernel is actually faster for finding the optimum. Clearer results with little haze residual can be found in Fig.~\ref{fig:parameter_gf}(d)-(e), which are concordant with above conclusion. In all the following experiments of this paper, we set the kernel radius of image guided filter as 32.

Figure~\ref{fig:parameter_rs} shows some results by setting the radiuses of local patch as 3, 5, 7, 9 and 11 when estimating $\eta_i^\lambda$ and $A_i^\lambda$. Other parameters are set as above accordingly. Fig.~\ref{fig:parameter_rs}(a)-(e) shows similar results. Larger patches are robust to noise and outliers, but also lead to a little more smooth results as well as more computation cost. As a trade-off, it is set as 5 in this paper.

\subsection{Comparisons with illumination correction methods}
\label{subsec:comparison_illumination_correction}
\begin{figure*}
\centering
\includegraphics[width=1\linewidth]{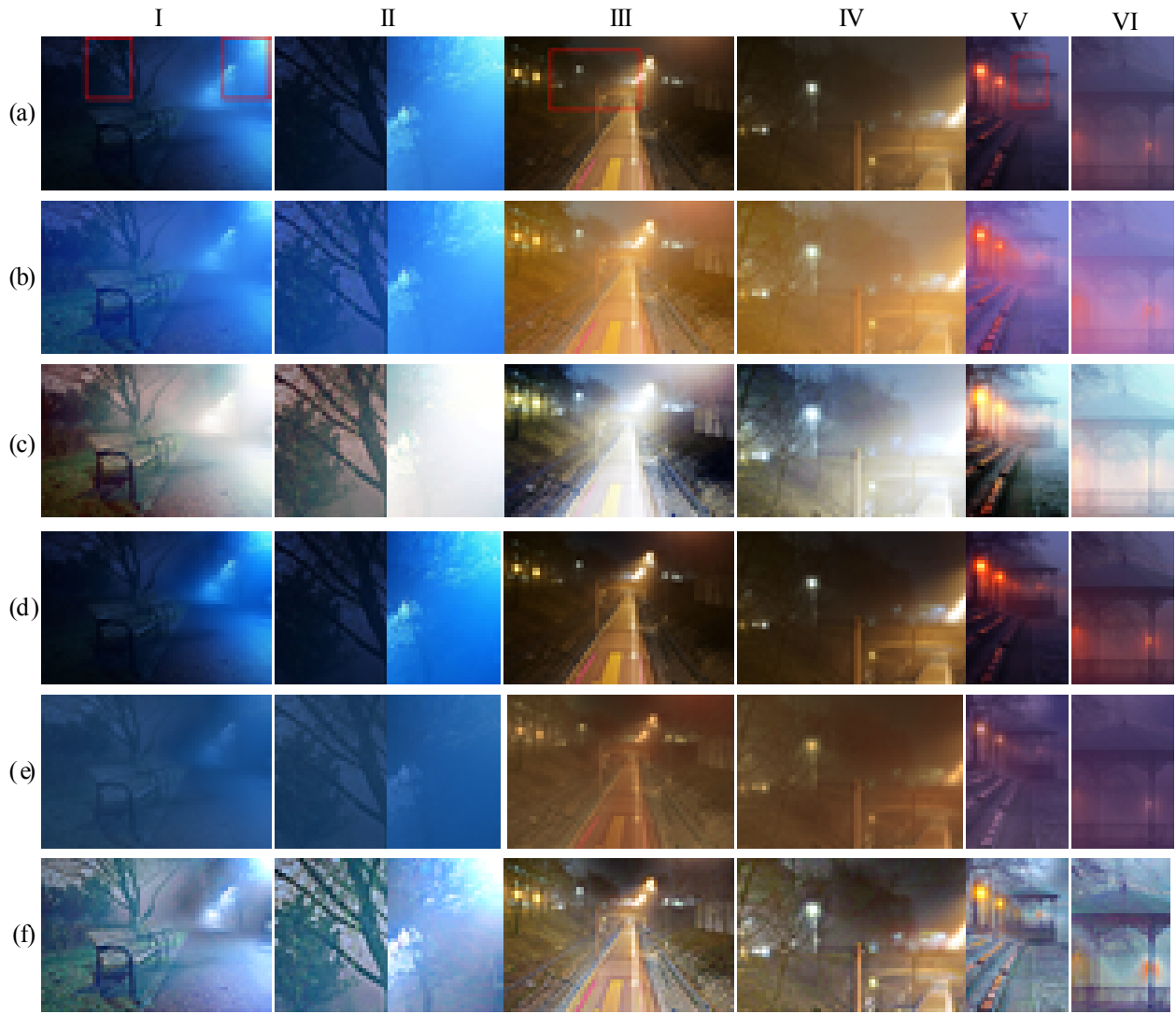}
\caption{(a) Original nighttime haze images. (b) Retinex results of \cite{SBH_2005_Elad}. (c) Results of Histogram equalization. (d) Results of He et al.'s method \cite{ECCV_2010_He}. (e) Results of Pei et al.'s method \cite{ICIP_2012_Pei}. (f) Results of the proposed algorithm.}
\label{fig:whole_comparisons}
\end{figure*}

In this experiment, we compared the proposed algorithm with other illumination correction methods, such as retinex using two bilateral filters \cite{SBH_2005_Elad} and histogram equalization (Heq), to evaluate their effectiveness on achieving illumination-balance result. Figure~\ref{fig:whole_comparisons}(a)-(c) and Figure~\ref{fig:whole_comparisons}(f) shows the original nighttime haze images, results of \cite{SBH_2005_Elad}, results of histogram equalization and results of the proposed algorithm, respectively. Images in column II, IV and VI are the close-up views of the red box regions in column I, III and V, respectively. It can be seen that the details in Fig. ~\ref{fig:whole_comparisons}(a) are hard to distinguish due to the non-uniform illumination and haze effects. Method in \cite{SBH_2005_Elad} can achieve illumination-balance results, but it cannot eliminate the color distortion and haze effects. Some parts of results shown in Fig.~\ref{fig:whole_comparisons}(b) are dominated by the colors of light sources and are in low contrast. Histogram equalization enhanced the overall contrast of the original images and achieved clearer results with more details than \cite{SBH_2005_Elad}. However, it also amplified the intensities in regions around light sources and reduced their contrast. Details in those part were not clear. In particular, when there is a wide gap of intensities within the whole image, i.e., some parts are very bright and some parts are very dark like column I and III of Fig.~\ref{fig:whole_comparisons}(a), this amplifying effect will be very strong. Besides, result for column I has color distortion, e.g., dull red in the left part of column II in Fig.~\ref{fig:whole_comparisons}(c). Compared with the above two methods, The proposed algorithm achieved the best results of more balanced illumination and less color distortion. Lots of clear details can be found in Fig.~\ref{fig:whole_comparisons}(f).

\subsection{Comparisons with dehazing methods}
\label{subsec:comparison_dehazing}
In this experiment, we compared the proposed algorithm with other dehazing methods, such as He et al.'s method \cite{ECCV_2010_He} and Pei et al.'s method \cite{ICIP_2012_Pei}, to evaluate their effectiveness on removing haze effects. Figure~\ref{fig:whole_comparisons}(d)-(f) shows results of He et al.'s method, Pei et al.'s method and the proposed algorithm, respectively. It can be seen that He et al.'s method only achieved a little better results than the original inputs. The reason is that the dark channel prior is ineffective for estimating transmission from images captured in nighttime environment. Pei et al.'s method achieved better results than He et al.'s method. More details can be found in their results. However, in the color transfer procedure of their method, it changed the color characteristics of the input nighttime haze image from a given target image by statistic correction. Since the different scene contents and imaging conditions, the global color transfer procedure led to a complete grayish scene. Some details were also lost in this procedure. Generally, this grayish result is not the expected illumination-balance one (Please see Fig.~\ref{fig:icip_tran}(e)). Consequently, Pei et al.'s method have not achieved satisfying results.

\begin{figure*}
\centering
\includegraphics[width=0.8\linewidth]{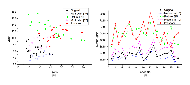}
\caption{(a) $\overline I$ and $\overline \sigma $ of results obtained by different methods on the 20 test images. (b) Visual measures of results obtained by different methods on the 20 test images.}
\label{fig:IxSigma}
\end{figure*}

\begin{figure*}
\centering
\includegraphics[width=0.8\linewidth]{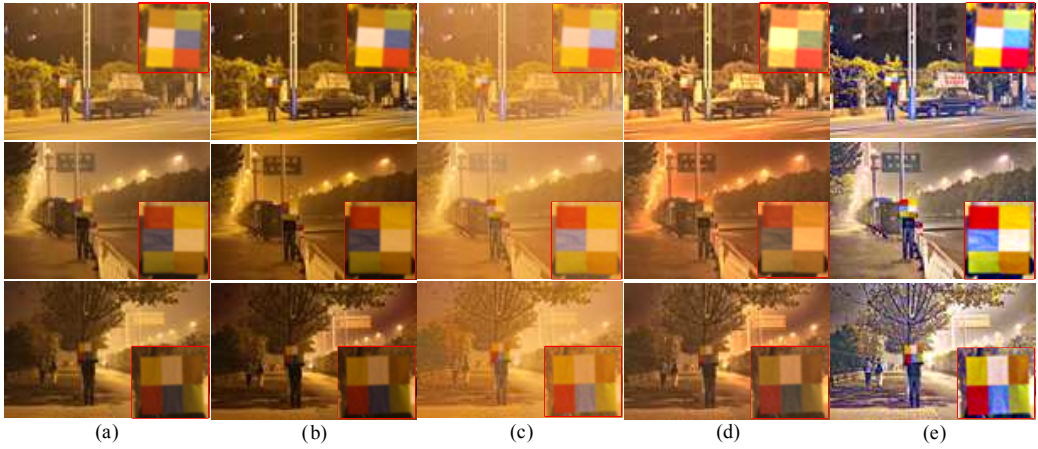}
\caption{Color rendition experiment on color set images captured in nighttime haze environment. (a) Original nighttime haze images. (b) Results of He et al.'s method \cite{ECCV_2010_He}. (c) Results of retinex method \cite{SBH_2005_Elad}. (d) Results of Pei et al.'s method \cite{ICIP_2012_Pei}. (e) Results of the proposed algorithm.}
\label{fig:color_rendition}
\end{figure*}

\begin{figure*}
\centering
\includegraphics[width=0.8\linewidth]{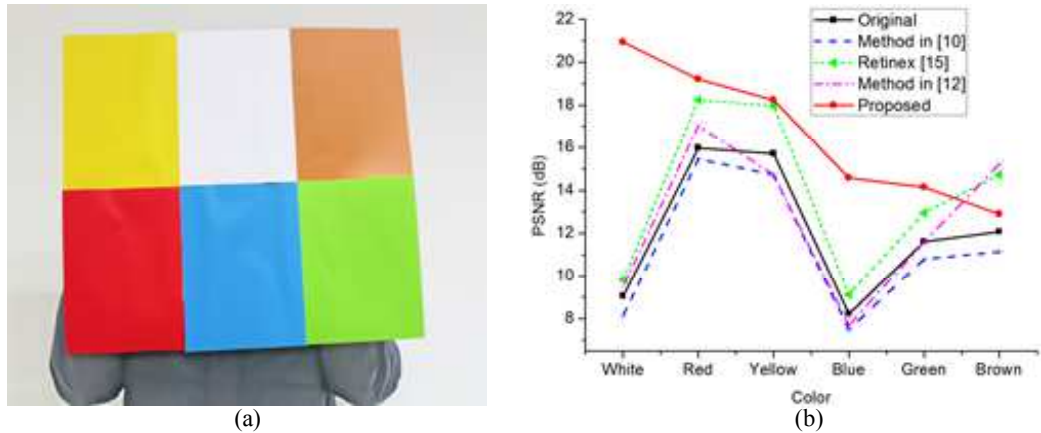}
\caption{(a) Ground truth color set captured in indoor environment with daylight lamp. (b) PSNRs of different methods on the six colors.}
\label{fig:colorpad_psnr}
\end{figure*}

\begin{figure*}
\centering
\includegraphics[width=0.9\linewidth]{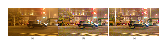}
\caption{(a) The $150^{th}$ frame of the test video sequence captured in nighttime haze environment. (b) Result of the proposed algorithm. (c) Denoising result of (b) by using BM3D denoising method \cite{TIP_2007_Dabov}.}
\label{fig:video_results}
\end{figure*}

Different from the statistical color transfer procedure in \cite{ICIP_2012_Pei}, the proposed algorithm is model-based and dose not require a target daytime image. It achieved illumination-balance result after estimating the model variables in the illumination compensation step and color correction step. Compared with the grayish result in \cite{ICIP_2012_Pei}, this result is more natural. Please compare Fig.~\ref{fig:icip_tran}(e) and Fig.~\ref{fig:flowchart}. It serves as a good input for the following dehazing step and led to the final illumination-balance and haze-free result. Generally, the proposed algorithm achieved better results than other methods mentioned in Section~\ref{subsec:comparison_illumination_correction} and Section~\ref{subsec:comparison_dehazing}. More results can be found on the webpage\footnote{\href{http://staff.ustc.edu.cn/\%7Eforrest/NighttimeDehazing.htm}{{http://staff.ustc.edu.cn/\%7Eforrest/NighttimeDehazing.htm}}}. The executable code is also available online.

\subsection{Statistical evaluation}
\label{subsec:statistical_evaluation}

In \cite{AS_2002_Jobson}, Jobson et al. propose a visual measure for automatically assessing the quality of visual representation. They found that visually optimized images are more tightly clustered about a single mean value and have much higher standard deviations. The visual measure in \cite{AS_2002_Jobson} is $\overline I  \times \overline \sigma $. An image is first divided into many non-overlapped patches, e.g., $50 \times 50$. Then, the mean value and standard deviation of intensities on each patches are calculated. Finally, $\overline I$ and $\overline \sigma $ can be calculated by averaging the mean values and standard deviations of those patches. Generally, the higher $\overline I  \times \overline \sigma $ is, the better the quality of an image is. When $\overline I$ lies in a range of [100, 200] and $\overline \sigma $  lies in a range of [40, 80], the image is considered as a visual good one. In this paper, we calculated the visual measures for results obtained by using different methods on the 20 test images. They are shown in Fig.~\ref{fig:IxSigma}. Generally, all the methods can enhance the intensity values of the original images except method in \cite{ECCV_2010_He}. The dehazing method obtained a result of lower intensities than the original one. Please check it from Eq.~\eqref{eq:daytime_model}. And few details (high variance regions) were recovered due to the directly using dark channel prior on nighttime haze images. Pei et al.'s method \cite{ICIP_2012_Pei} achieved a little higher $\overline I$ and $\overline \sigma $, thus a little higher visual measure than the original images. Significant enhancement were achieved by the retinex method in \cite{SBH_2005_Elad} and the proposed algorithm. However, there were different trends of their results. In Fig.~\ref{fig:IxSigma}(a), each pair of $\overline \sigma $ and $\overline I$ were plotted as a point in the plane. The points of the retinex method in \cite{SBH_2005_Elad} lied in a broad range, but the points of the proposed algorithm tended to be clustered around $\left( {45,110} \right)$. According to \cite{AS_2002_Jobson}, the proposed algorithm achieved visual good results. It is consistent with the visual results in Fig.~\ref{fig:whole_comparisons}. The visual measures of results from different methods were plotted in Fig.~\ref{fig:IxSigma}(b). The average values are $1.7627 \times 10^3$, $1.3811 \times 10^3$, $4.9499 \times 10^3$, $2.5234 \times 10^3$ and $4.9902 \times 10^3$, for the original images, method in \cite{ECCV_2010_He}, method \cite{SBH_2005_Elad}, Pei et al.'s method \cite{ICIP_2012_Pei} and the proposed algorithm, respectively.

\subsection{Evaluation on color rendition}
\label{subsec:color_rendition}
Moreover, we also conducted an experiment to evaluate the color rendition abilities of different methods. We used a color set including colors of yellow, white, brown, red, blue and green. The images captured in nighttime haze environment and the corresponding results of different methods are shown in Fig.~\ref{fig:color_rendition}. The groud truth is shown in Fig.~\ref{fig:colorpad_psnr}(a), which is captured in indoor environment with daylight lamp. We calculated the PSNRs of different results for the above six colors. The results were plotted in Fig.~\ref{fig:colorpad_psnr}(b). It can be seen that the proposed algorithm achieved better result than other methods, especially for white and blue. Only PNSR for brown was a little lower than  \cite{SBH_2005_Elad} and \cite{ICIP_2012_Pei}. In Fig.~\ref{fig:color_rendition}(a)-(d), there are obvious color distortions. The average PSNRs are 12.11dB, 11.30dB, 13.81dB, 12.63dB and 16.67dB for the original images, method \cite{ECCV_2010_He}, method \cite{SBH_2005_Elad}, method \cite{ICIP_2012_Pei} and the proposed algorithm, respectively. The proposed algorithm shows better color rendition ability than other methods for nighttime images enhancement.

\subsection{Video results in nighttime haze environment}
\label{subsec:video_results}
We also evaluated our algorithm on a video sequence captured at a crossroad in a foggy night. It consists of 635 frames and lasts 42 seconds at a speed of 15fps. Figure~\ref{fig:video_results}(a) shows the $150^{th}$ frame of the original sequence, and Fig.~\ref{fig:video_results}(b) shows the dehazing result of our algorithm. It can be seen that our algorithm can achieve a clear result with little color distortion. Please compare the regions like marks on the road, trees, cars parked on the other side of the road and the building.

Noise is a common problem for images captured in nighttime. And image enhancement algorithm often amplifies these noises. Recently, different efficient denoising algorithm have been proposed for noisy images and video sequences \cite{CVPR_2005_Buades, MMS_2006_Buades, SPARS_2009_Dabov, TIP_2007_Dabov}. Since image/video denosing is another hot research area, we limit our discussions on it and leave it as future work. Actually, we can include the BM3D algorithm presented in \cite{TIP_2007_Dabov} as a post-processing step in the proposed algorithm. Figure~\ref{fig:video_results}(c) shows the denoising result of Fig.~\ref{fig:video_results}(b). The complete test video and corresponding dehazing results with/without denoising step can be download from \href{http://pan.baidu.com/s/1zYHDO}{http://pan.baidu.com/s/1zYHDO}.

\subsection{Quantitative results on synthetic nighttime hazy images}
\label{subsec:synthetic_images}

\begin{figure*}
\centering
\includegraphics[width=0.9\linewidth]{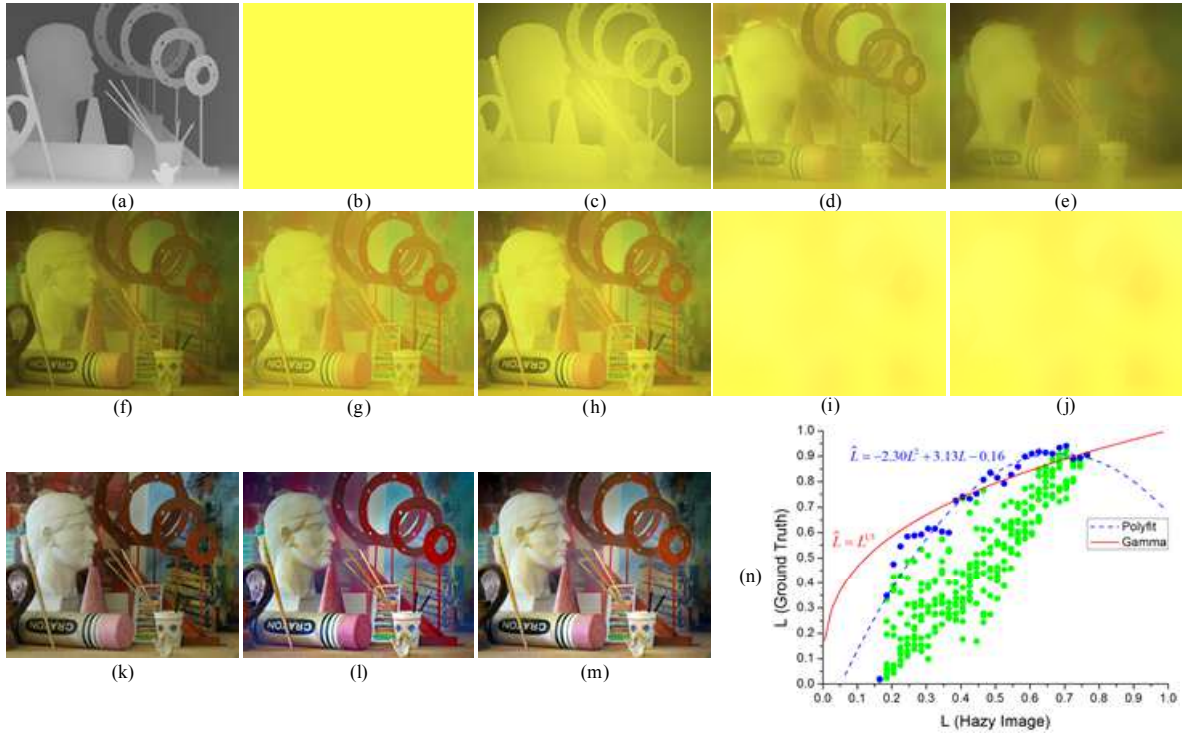}
\caption{Synthetic results on test image of ``Art''. (a) Disparity map. (b) Color of light source, i.e., $\eta ^\lambda$. (c) Illumination image. (d) Color of environmental light, i.e., $\sigma ^\lambda$. (e) Environmental light image. (f) Synthetic nighttime hazy image. (g)-(h) Illumination compensation result of gamma correction and polynomial fitting. (i)-(j) Estimates of $\eta ^\lambda$ on (g)-(h). (k) Ground truth. (l)-(m) Final dehazing results by using gamma correction and polynomial fitting in the proposed algorithm, respectively. (n) Gamma curve and polynomial fitting curve about points on the upper bound.}
\label{fig:synthesisExperiments}
\end{figure*}

In this section, we used images from Middlebury 2005 dataset\footnote{\href{http://vision.middlebury.edu/stereo/data/scenes2005/}{http://vision.middlebury.edu/stereo/data/scenes2005/}} to synthesize the nighttime hazy images, and quantitatively evaluated the performance of the proposed algorithm. By referring the synthetic method in \cite{CVPR_2014_Tang}, we calculated the transmission map as $t = 0.8d$, where $d$ is the normalized disparity map. According to the binocular triangulation measurement principle, we calculated the coordinate of each pixel in the world coordinate system. Then, we assumed that the only light source is located at position $(0,0,0)$, i.e., in front of the center part of the scene. So we calculated the illumination value of a scene point by using a negative exponential form as $L = \exp \left( { - \beta  \times dis} \right)$, where $\beta$ is a parameter and $dis$ represent the normalized distance between a scene point and the light source. Since $\beta$ is small, we used its first order Taylor series expansion instead, i.e., $L = 1 - \beta  \times dis$. In the following experiment, $\beta$ was set to 0.8. We calculated the environmental light according to Eq.~\eqref{eq:environmental_light}. Instead of averaging the scattered incident light and reflected light, we applied an image guided filtering process on them with a large regularization parameter, i.e., 0.1.  The parameter $\alpha$ was set to 0.5, and the radius of local patch was set to 16. As yellow is a common color in artificial light sources such as road lamp, so we set the color of the light source in our synthetic experiment as $\left( {1,1,0.3} \right)$, i.e., $\eta {\rm{ = }}\left( {1,1,0.3} \right)$. In addition, the original clear image in the dataset was used as the reflectance $R$. Finally, we generated the nighttime hazy image according to Eq.~\eqref{eq:our_model} and Eq.~\eqref{eq:environmental_light}. An example is shown in Fig.~\ref{fig:synthesisExperiments}. Figure~\ref{fig:synthesisExperiments}(c) and (e) show the synthetic illumination image and environmental light image, respectively. And Fig.~\ref{fig:synthesisExperiments}(f) shows the final synthetic nighttime hazy image.

\begin{figure*}
\centering
\includegraphics[width=0.9\linewidth]{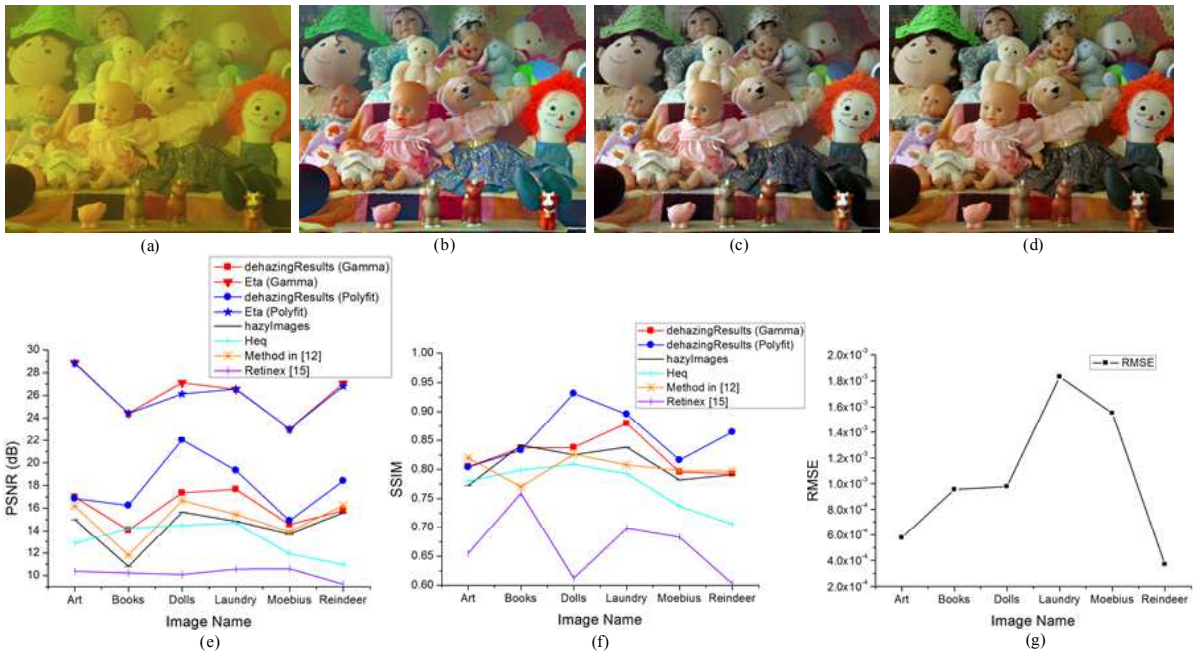}
\caption{(a) Synthetic nighttime hazy image of ``Dolls''. (b)-(c) Final dehazing reuslts by using gamma correction and polynomial fitting in the proposed algorithm, respectively. (d) Ground truth. (e) PSNR indices of dehzing reuslts and estimates of $\eta ^\lambda$. (f) SSIM indices of dehazing results. (e) RMSE between $\eta ^\lambda$ and $\max \left\{ {{\eta ^\lambda },{\sigma ^\lambda }} \right\}$.}
\label{fig:synthesisExperimentsMore}
\end{figure*}

Besides the gamma correction in the illumination compensation step of the proposed algorithm, we can also compensate the illumination by using polynomial fitting since we have the ground truth in this synthetic experiment. Figure~\ref{fig:synthesisExperiments}(n) shows the polynomial fitting curve about points on the upper bound. The horizontal axis and vertical axis denote the illumination values of the synthetic nighttime hazy image and original clear image, respectively. Figure~\ref{fig:synthesisExperiments}(g)-(h) show the illumination compensation results by using gamma correction and polynomial fitting, respectively. The illumination of gamma correction result seems more balanced, but the haze effect is amplified, e.g., haze in Fig.~\ref{fig:synthesisExperiments}(g) is thicker. Figure~\ref{fig:synthesisExperiments}(i)-(j) are the subsequent estimates of $\eta ^\lambda$. Both results are similar to each other and close to the ground truth (Fig.~\ref{fig:synthesisExperiments}(b)). Figure~\ref{fig:synthesisExperiments}(l)-(m) are the final corresponding dehazing results. Result which corresponds to gamma correction is a little over-saturation. It is affected by the amplified haze effect. Result which corresponds to polynomial fitting is closer to the ground truth (Fig.~\ref{fig:synthesisExperiments}(k)). But its illumination is not so balanced as Fig.~\ref{fig:synthesisExperiments}(l), e.g., up-left and up-right corners. Another example can be found in Fig.~\ref{fig:synthesisExperimentsMore}(a)-(d). PSNR and SSIM indices of all results are shown in Figure~\ref{fig:synthesisExperimentsMore}(e)-(f). Both illumination compensation technique in the proposed algorithm achieved higher gains than the nighttime hazy image. And PSNR indices of both estimates of $\eta ^\lambda$ are very high. Besides, we calculated the RMSE between the synthetic $\eta ^\lambda$ and $\max \left\{ {{\eta ^\lambda },{\sigma ^\lambda }} \right\}$. The indices (Fig.~\ref{fig:synthesisExperimentsMore}(g)) show that their differences are negligible. It supports the conclusion in Eq.~\eqref{eq:environmental_light_inequality}, i.e., $\sigma _i^\lambda  \le \eta _i^\lambda $.

\subsection{Computational complexity analysis}
\label{subsec:complexity_analysis}
\begin{figure}
\centering
\includegraphics[width=0.9\linewidth]{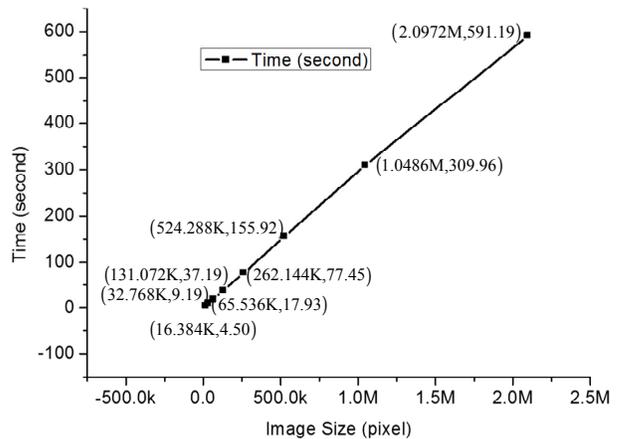}
\caption{Computation time v.s. image size.}
\label{fig:time}
\end{figure}

The computational cost mainly concentrates on five aspects, i.e., estimation about $L_i$, $\widehat{R_i^\lambda }$, $\eta_i^\lambda$, $t_i$ and $A_i^\lambda$. The estimation about the first two variables are implemented by using image guided filter which has an $O\left( N \right)$ time exact algorithm \cite{ECCV_2010_He}. The estimation about $\eta_i^\lambda$ based on some max operations on local patches, which is very similar to the estimation about transmission map in \cite{ECCV_2010_He}. The following refinement about the above raw estimate of $\eta_i^\lambda$ is also implemented by using image guided filter. The estimation about $t_i$ is same as \cite{ECCV_2010_He} based on the dark channel prior. And it is also refined by using image guided filter. The estimation about $A_i^\lambda$ in this paper is same as \cite{ECCV_2010_He} but in a local neighborhood rather than in the whole image when estimating $A_i$ in \cite{ECCV_2010_He}. Consequently, the proposed algorithm also has a linear computational complexity with regard to the image size. We implemented our algorithm using Matlab on a Laptop with Intel Core i5 and memory of 8GB. As an example, we calculated the computational times of test images of different sizes, e.g., $128 \times 128$, $128 \times 256$, $256 \times 256$, $256 \times 512$, $512 \times 512$, $512 \times 1024$, $1024 \times 1024$ and $1024 \times 2048$. Figure.~\ref{fig:time} shows the results.

We did not adopt any advanced acceleration techniques such as CPU-based/GPU-based parallel computing in the above implementation. Indeed, different acceleration techniques have been proposed for image dehazing recently \cite{JRTIP_2012_Zhang, VC_2012_Xiao, Fujitsu_2014_Tan, PG_2010_Lv, ICDIP_2013_Jin}. We leave it as future work to realize a faster implementation of the proposed algorithm.

\subsection{Limitation and discussions}
\label{subsec:limitation}
Since there are many unknown variables in the new imaging model (Eq.~\eqref{eq:our_model}) and the imaging conditions in nighttime haze environment are diverse, the proposed sequential solution based on some simple priors may not achieve satisfying solutions for all possible cases. For instance, color distortions can be observed in some parts of the trees in Fig.~\ref{fig:color_rendition}(e). Due to the low illuminations and the lack of color information in these regions, the estimates of $\eta_i^\lambda$ may be incorrect. It will lead color distortions on the final dehazing result. In the future work, it seems feasible by using more prior knowledge as well as some interactive techniques to overcome this problem.

\section{Conclusion}
In this paper, we propose an efficient algorithm to increase the visibility of the nighttime images. Based on the analysis of physical properties, we present a new imaging model for nighttime haze environment. This new model takes into account both the non-uniform illumination from artificial light sources and the scattering and attenuation effects of haze. Based on simple assumptions, the variables in the model can be efficiently estimated through a three-step sequential solution including illumination compensation, color correction and dehazing. Experimental results on real/synthetic nighttime hazy images show that the proposed algorithm can achieve illumination-balance and haze-free results. Meanwhile, it also has good color rendition ability and low computational cost. The future work may concentrate on realizing a faster implementation of the proposed algorithm and developing efficient dehazing algorithm of videos by using the temporal redundancy information.

\section*{Acknowledgment}
This work is supported by the National Science and Technology Major Project of the Ministry of Science and Technology of China (No.2012GB102007) and NSFC (No.61472380). Thanks for KANG Kai's assistance and all the owners of the test images from Flickr.

\appendix
Since we assume that $L_i$ and $\eta_i^\lambda$ are piecewise smooth, so we can rewrite $L_j$ and $\eta_j^\lambda$ in Eq.~\eqref{eq:environmental_light} as ${L_j} = {L_i} + \varepsilon _j^L,\forall j \in {\Omega _i}$ and $\eta _j^\lambda  = \eta _i^\lambda  + \varepsilon _j^\eta ,\forall j \in {\Omega _i}$, where $\varepsilon _j^L$ and $\varepsilon _j^\eta $ are two small quantities relative to $L_i$ and $\eta_i^\lambda$, respectively. Looking back at Eq.~\eqref{eq:environmental_light}, we have:
\begin{equation}
\begin{array}{l}
 B_i^\lambda  \buildrel \Delta \over = {L_i}\sigma _i^\lambda \\
 \quad \ \ \le \frac{1}{{\left| \Omega  \right|}}\sum\limits_{j \in \Omega_i} {{L_j}\eta _j^\lambda }  \\
 \quad \ \ {\rm{ = }}\frac{1}{{\left| \Omega  \right|}}\sum\limits_{j \in \Omega_i} {\left( {{L_i} + \varepsilon _j^L} \right)\left( {\eta _i^\lambda  + \varepsilon _j^\eta } \right)}  \\
 \quad \ \ = \frac{1}{{\left| \Omega  \right|}}\sum\limits_{j \in \Omega_i} {{L_i}\eta _i^\lambda }  + {L_i}\left( {\frac{1}{{\left| \Omega  \right|}}\sum\limits_{j \in \Omega_i} {\varepsilon _j^\eta } } \right) \\
 \qquad \ \ + \eta _i^\lambda \left( {\frac{1}{{\left| \Omega  \right|}}\sum\limits_{j \in \Omega_i} {\varepsilon _j^L} } \right) + \frac{1}{{\left| \Omega  \right|}}\sum\limits_{j \in \Omega_i} {\varepsilon _j^L\varepsilon _j^\eta }  \\
 \quad \ \ \approx {L_i}\eta _i^\lambda  \\
 \end{array}.
\label{eq:environmental_light_inequality}
\end{equation}
Because $\varepsilon _j^L$ is offset by each other in the local neighborhood $\Omega_i$ (so is $\varepsilon _j^\eta $), and the last term is a relative small quantity, so the last equality in Eq.~\eqref{eq:environmental_light_inequality} approximately holds. It implies that $\sigma _i^\lambda$ is smaller than $\eta _i^\lambda $, i.e., $\sigma _i^\lambda  \le \eta _i^\lambda $.

Similarly, we can rewrite $R_j^\lambda$, $\sigma _j^\lambda$ and $t_j$ as $R_j^\lambda  = R_i^\lambda  + \varepsilon _j^R$, $\sigma _j^\lambda  = \sigma _i^\lambda  + \varepsilon _j^\sigma$ and $t_j = t_i + \varepsilon _j^t ,\forall j \in {\Omega _i}$, where $\varepsilon _j^R $, $\varepsilon _j^\sigma$ and $\varepsilon _j^t$ are small quantities relative to $R_i^\lambda$, $\sigma_i^\lambda$ and $t_i$, respectively. Then, we have:
\begin{equation}
\begin{array}{l}
 \left| {\widehat{R_i^\lambda }{\rm{ - }}\widehat{R_j^\lambda }} \right| = \left| {\eta _i^\lambda R_i^\lambda {t_i} + \sigma _i^\lambda \left( {1 - {t_i}} \right){\rm{ - }}\eta _j^\lambda R_j^\lambda {t_j} - \sigma _j^\lambda \left( {1 - {t_j}} \right)} \right| \\
 \qquad \qquad \ = \left| \begin{array}{l}
 \eta _i^\lambda R_i^\lambda {t_i} - \left( {\eta _i^\lambda  + \varepsilon _j^\eta } \right)\left( {R_i^\lambda  + \varepsilon _j^R} \right)\left( {{t_i} + \varepsilon _j^t} \right) \\
  + \sigma _i^\lambda \left( {1 - {t_i}} \right) - \left( {\sigma _i^\lambda  + \varepsilon _j^\sigma } \right)\left( {1 - {t_i} - \varepsilon _j^t} \right) \\
 \end{array} \right| \\
 \qquad \qquad \ \le \left| {\varepsilon _j^\eta } \right| + \left| {\varepsilon _j^R} \right|{\rm{ + }}\left| {\varepsilon _j^t} \right| + \left| {\varepsilon _j^\eta \varepsilon _j^R} \right| + \left| {\varepsilon _j^\eta \varepsilon _j^t} \right| + \left| {\varepsilon _j^R\varepsilon _j^t} \right|\\
  \qquad \qquad \qquad + \left| {\varepsilon _j^\eta \varepsilon _j^R\varepsilon _j^t} \right| + 2\left| {\varepsilon _j^\sigma } \right| + \left| {\varepsilon _j^t} \right| + \left| {\varepsilon _j^\sigma \varepsilon _j^t} \right| \\
 \end{array},
\label{eq:surrogate_reflectance_inequality}
\end{equation}
The last inequality holds because $\eta_i^\lambda$, $R_i^\lambda$, $\sigma_i^\lambda$ and $t_i$ all lie in the range of [0,1]. Eq.~\eqref{eq:surrogate_reflectance_inequality} shows that $\widehat{R_i^\lambda }$ is piecewise continuous.


%





\ifCLASSOPTIONcaptionsoff
  \newpage
\fi

\end{document}